\pdfoutput=1

\documentclass[11pt]{article}

\usepackage[]{acl}

\usepackage{times}
\usepackage{latexsym}
\usepackage{multirow}
\usepackage{booktabs}
\usepackage{graphicx}
\usepackage{amssymb}
\usepackage{makecell}
\usepackage{amsmath, bm}
\usepackage{longtable}
\usepackage{subfigure}
\usepackage{caption}
\usepackage{enumerate}
\usepackage{enumitem}

\usepackage[T1]{fontenc}

\usepackage[utf8]{inputenc}

\usepackage{microtype}
 
\title{Few-shot Named Entity Recognition with Self-describing Networks}

\author{
  Jiawei Chen${}^{1,3,}$\thanks{~ Equally Contribution.},
  Qing Liu${}^{1,3,}$\footnotemark[1],
  Hongyu Lin${}^{1,}$\thanks{~ Corresponding authors.},
  Xianpei Han${}^{1,2,4,}$\footnotemark[2],
  Le Sun${}^{1,2}$
  \\
  ${}^{1}$Chinese Information Processing Laboratory ~
  ${}^{2}$State Key Laboratory of Computer Science \\
  Institute of Software, Chinese Academy of Sciences, Beijing, China\\
  ${}^{3}$University of Chinese Academy of Sciences, Beijing, China \\
  ${}^{4}$Beijing Academy of Artificial Intelligence, Beijing, China \\
  {\tt \{jiawei2020,liuqing2020,hongyu,xianpei,sunle\}@iscas.ac.cn} \\
}

\begin{document}
\maketitle
\begin{abstract}
Few-shot NER needs to effectively capture information from limited instances and transfer useful knowledge from external resources. In this paper, we propose a self-describing mechanism for few-shot NER, which can effectively leverage illustrative instances and precisely transfer knowledge from external resources by describing both entity types and mentions using a universal concept set. Specifically, we design \textit{Self-describing Networks} (SDNet), a Seq2Seq generation model which can universally describe mentions using concepts, automatically map novel entity types to concepts, and adaptively recognize entities on-demand. We pre-train SDNet with large-scale corpus, and conduct experiments on 8 benchmarks from different domains. Experiments show that SDNet achieves competitive performances on all benchmarks and achieves the new state-of-the-art on 6 benchmarks, which demonstrates its effectiveness and robustness.

\end{abstract}

\section{Introduction}
Few-shot named entity recognition (FS-NER) aims to identify entity mentions corresponding to new entity types (i.e., novel types) with only a few illustrative examples. FS-NER is a promising technique for open-domain NER which contains various unforeseen types and very limited examples and therefore has attached great attention in recent years~\citep{huang,wang}.

\begin{figure}[t!]
\centering 
\setlength{\belowcaptionskip}{-0.4cm}
\includegraphics[width=0.49\textwidth]{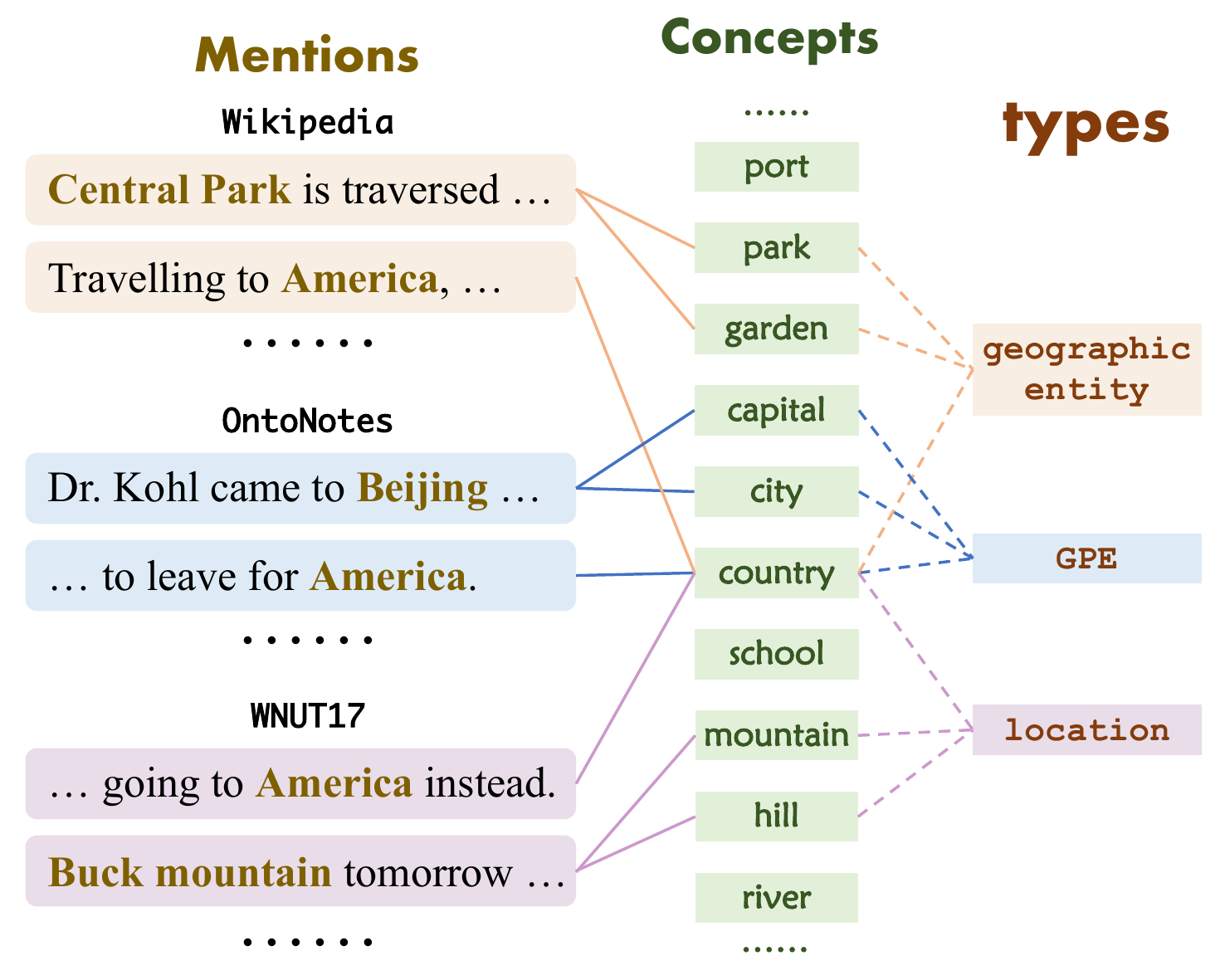}
\caption{Examples of concept description. Wikipedia, OntoNotes, WNUT17 can transfer knowledge to each other by describing mentions and types using a universal concept set.}
\label{Fig.introduction}
\end{figure}

The main challenge of FS-NER is how to accurately model the semantics of unforeseen entity types using only a few illustrative examples. To achieve this, FS-NER needs to effectively capture information in few-shot examples, meanwhile exploiting and transferring useful knowledge from external resources. Unfortunately, information entailed in illustrative examples is very limited, i.e., the \textbf{limited information challenge}. And external knowledge usually doesn't directly match with the new task because it may contain irrelevant, heterogeneous, or even conflicting knowledge~\citep{DBLP:conf/acl/BeryozkinDGHS19,DBLP:journals/corr/abs-2004-05140} -- which we refer as \textbf{knowledge mismatch challenge}. For example, the schemas in Wikipedia, OntoNotes~\citep{ontonotes} and WNUT17~\citep{wnut} are conflicting, where ``America'' is \texttt{geographic entity} in Wikipedia, \texttt{GPE} in OntoNotes, and \texttt{location} in WNUT17. Such a knowledge mismatch problem makes it unsuitable to directly transfer external knowledge to downstream tasks. Consequently, how to sufficiently leverage limited few-shot examples and precisely transfer external knowledge are the critical challenges for FS-NER.

To this end, this paper proposes a self-describing mechanism for FS-NER. The main idea behind self-describing mechanism is that all entity types can be described using the same set of concepts, and the mapping between types and concepts can be universally modeled and learned. In this way, the knowledge mismatch challenge can be resolved by uniformly describing different entity types using the same concept set. For example, in Figure~\ref{Fig.introduction} the types in different schemas are mapped to the same concept set \{\textit{park}, \textit{garden}, \textit{country}, ...\}, therefore the knowledge in different sources can be universally described and transferred. Furthermore, because the concept mapping is universal, the few examples are only used to construct the mapping between novel types and concepts, the limited information problem can be effectively addressed.
 
Based on the above idea, we propose Self-describing Networks -- SDNet, a Seq2Seq generation network which can universally describe mentions using concepts, automatically map novel entity types to concepts, and adaptively recognize entities on-demand. Specifically, to capture the semantics of a mention, SDNet generates a set of universal concepts as its description. For example, generate \{\textit{capital, city}\} for “Dr. Kohl came to [Beijing], ...”. To map entity types to concepts, SDNet generates and fuses the concept description of the mentions with the same entity type. For example, map \texttt{GPE} to \{\textit{country}, \textit{capital}, \textit{city}\} using its mentions  “Beijing” and “America”. To recognize entity, SDNet directly generates all entities in a sentence via a concept-enriched prefix prompt, which contains the target entity types and their concept descriptions. For example, recognizing entity in “France is beautiful.”  by generating“France is GPE” using prefix prompt “\texttt{[EG]} \texttt{GPE}: \{\textit{country}, \textit{capital}, \textit{city}\}”. Because the concept set is universal, we pre-train SDNet on large-scale, easily accessible web resources. Concretely, we collect a pre-training dataset which contains 56M sentences with more than 31K concepts by leveraging the links from Wikipedia anchor words to the Wikidata items. 

By projecting both mentions and entity types to a universal concept space, SDNet can effectively enrich entity types to resolve the limited information problem, universally represent different schemas to resolve the knowledge mismatch problem, and can be effectively pre-trained in a unified way. Moreover, all the above tasks are modeled in a single generation model by using prefix prompt mechanism to distinguish different tasks, which makes the model controllable, universal and can be continuously trained.

We conduct experiments on 8 few-shot NER benchmarks with different domains. Experiments show that SDNet leads to very competitive performance and achieves the new state-of-the-art on 6 of these benchmarks.\footnote{Our source codes are openly available at \url{https://github.com/chen700564/sdnet}}

Generally speaking, the contributions of this paper are:
\begin{itemize}[leftmargin=0.6cm,topsep=0.1cm]
\setlength{\itemsep}{0cm}
\setlength{\parskip}{0.1cm}
\item We propose a self-describing mechanism for FS-NER, which can effectively resolve the limited information challenge and the knowledge mismatch challenge by describing both entity types and mentions using a universal concept set.
\item We propose Self-describing Networks -- SDNet, a Seq2Seq generation network which can universally describe mentions using concepts, automatically map novel entity types to concepts, and adaptively recognize entities on-demand.
\item We pre-train SDNet on the large-scale open dataset, which provides a universal knowledge for few-shot NER and can benefit many future NER studies.
\end{itemize}

\section{Related Work}
To deal with the limited information challenge, current FS-NER studies mostly focus on leveraging external knowledge, many knowledge resources are used: 
1) PLMs. Early FS-NER studies~\citep{tong-etal-2021-learning,wang} mainly use PLMs for better encoding. And prompt-based NER formulation is proposed to exploit the PLMs' knowledge more effectively~\citep{DBLP:conf/emnlp/XinZH0S18,DBLP:conf/naacl/ObeidatFST19,DBLP:conf/acl/DaiSW20, DBLP:journals/corr/abs-2108-10604,uner,DBLP:conf/emnlp/LiuLXH0W21,DBLP:conf/acl/CuiWLYZ21,DBLP:journals/corr/abs-2109-13532, DBLP:journals/corr/abs-2110-08454}. 
2) Existing annotation datasets. These studies~\citep{fritzler2019few,DBLP:conf/acl/HouCLZLLL20,DBLP:conf/emnlp/YangK20,li2020few,li2020metaner,tong-etal-2021-learning,das2021container} focus on reusing annotations in existing datasets, and the annotations can be used to pre-train NER models.
3) Distantly annotated datasets. Some works~\citep{mengge-etal-2020-coarse,huang,jiang-etal-2021-named} try to automatically construct NER datasets via distant supervision, but which often suffer from the partially-labeled~\citep{partial_anno2018, partial_anno2019, DBLP:conf/acl/PengXZFH19} and noise label~\citep{DBLP:conf/emnlp/ShangLGRR018,DBLP:conf/acl/PengXZFH19,DBLP:conf/aaai/ZhangLH0LWY21,ZhangLH020} problem.

To deal with the knowledge mismatch problem, \citet{kim-etal-2015-new,DBLP:conf/cvpr/ReedALS16,DBLP:conf/cvpr/QiaoLSH16,DBLP:conf/cvpr/XianC0SA19,DBLP:conf/acl/HouCLZLLL20} employ label project methods which project labels in different schemas. \citet{rei-sogaard-2018-zero,li-etal-2020-unified,wang,DBLP:conf/acl/Aly0M20} enrich the semantics of labels using manually label descriptions. \citet{DBLP:conf/acl/BeryozkinDGHS19, DBLP:journals/corr/abs-2004-05140} merge the labels in different schemas into the same taxonomy for knowledge sharing. And \citet{jiang-etal-2021-named} relabels the external noisy datasets using current labels. Compared with these methods, we resolve the knowledge mismatch problem by mapping all entity types to a universal concept set, and the concept mapping and target entities are automatically generated using a self-describing networks.

\section{Self-describing Networks for FS-NER}
\begin{figure*}[tbh!]
\centering
\includegraphics[width=1.0\textwidth]{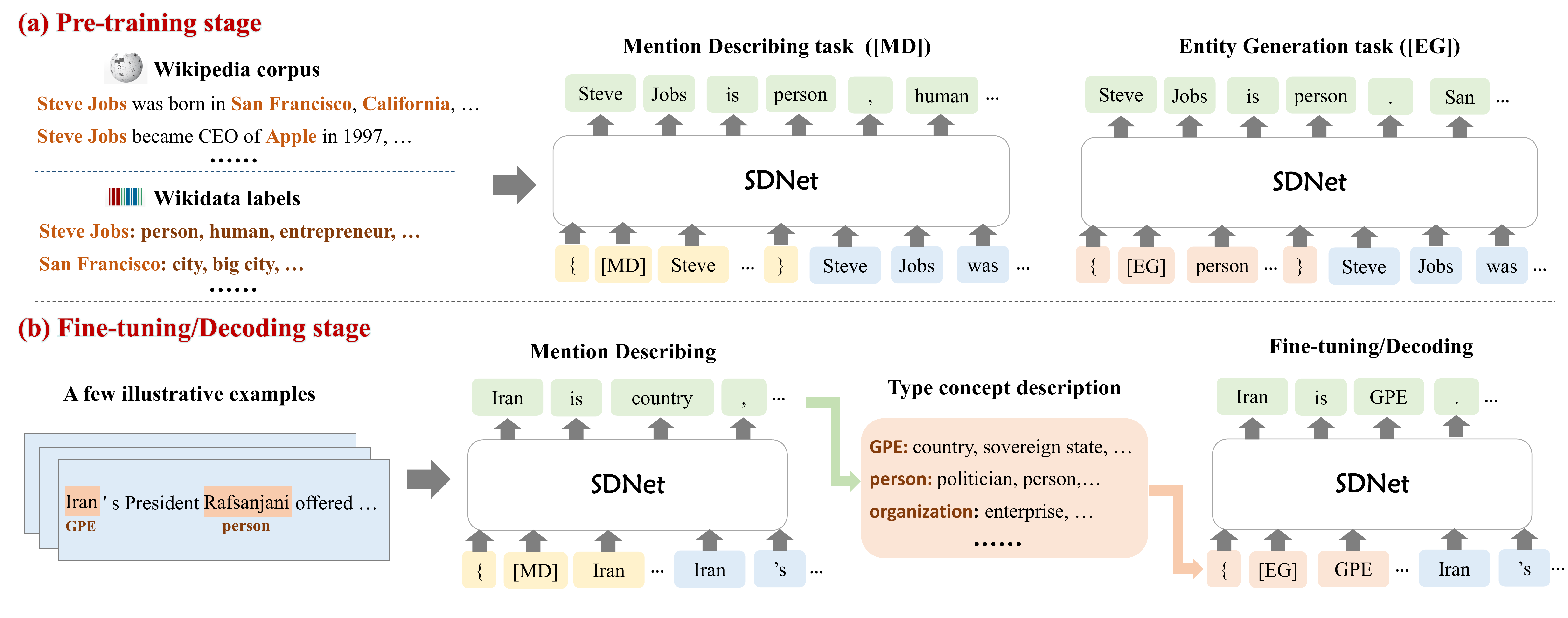}
\caption{Overview of the process of SDNet. The upper part is the pre-training stage, and the lower part is the fine-tuning/decoding stage. In pre-training stage, the external data is used to jointly train mention describing and entity generation tasks. In fine-tuning/decoding stage, SDNet first conducts mention describing to summarize type concept descriptions, and then conducts entity generation based on the generated descriptions.}
\label{Fig.framework}
\end{figure*}

In this section, we describe how to build few-shot entity recognizers and recognize entities using Self-describing networks. Figure~\ref{Fig.framework} (b) shows the entire procedure. Specifically, SDNet is a Seq2Seq network which performs two generation tasks successively 1) \textbf{\textit{Mention describing}}, which generates the concept descriptions of mentions; 2) \textbf{\textit{Entity generation}}, which adaptively generates entity mentions corresponding to desirable novel type one by one. Using SDNet, NER can be directly performed through the entity generation process by putting type descriptions into its prompt. Given a novel type, its type description is built through mention describing upon its illustrative instances. In the following, we will first introduce SDNet, then describe how to construct type descriptions and build few-shot entity recognizers.

\subsection{Self-describing Networks}\label{overall}
\begin{figure}[t!]
\setlength{\belowcaptionskip}{-0.4cm}
\setlength{\abovecaptionskip}{0.2cm}
\centering 
\includegraphics[width=0.49\textwidth]{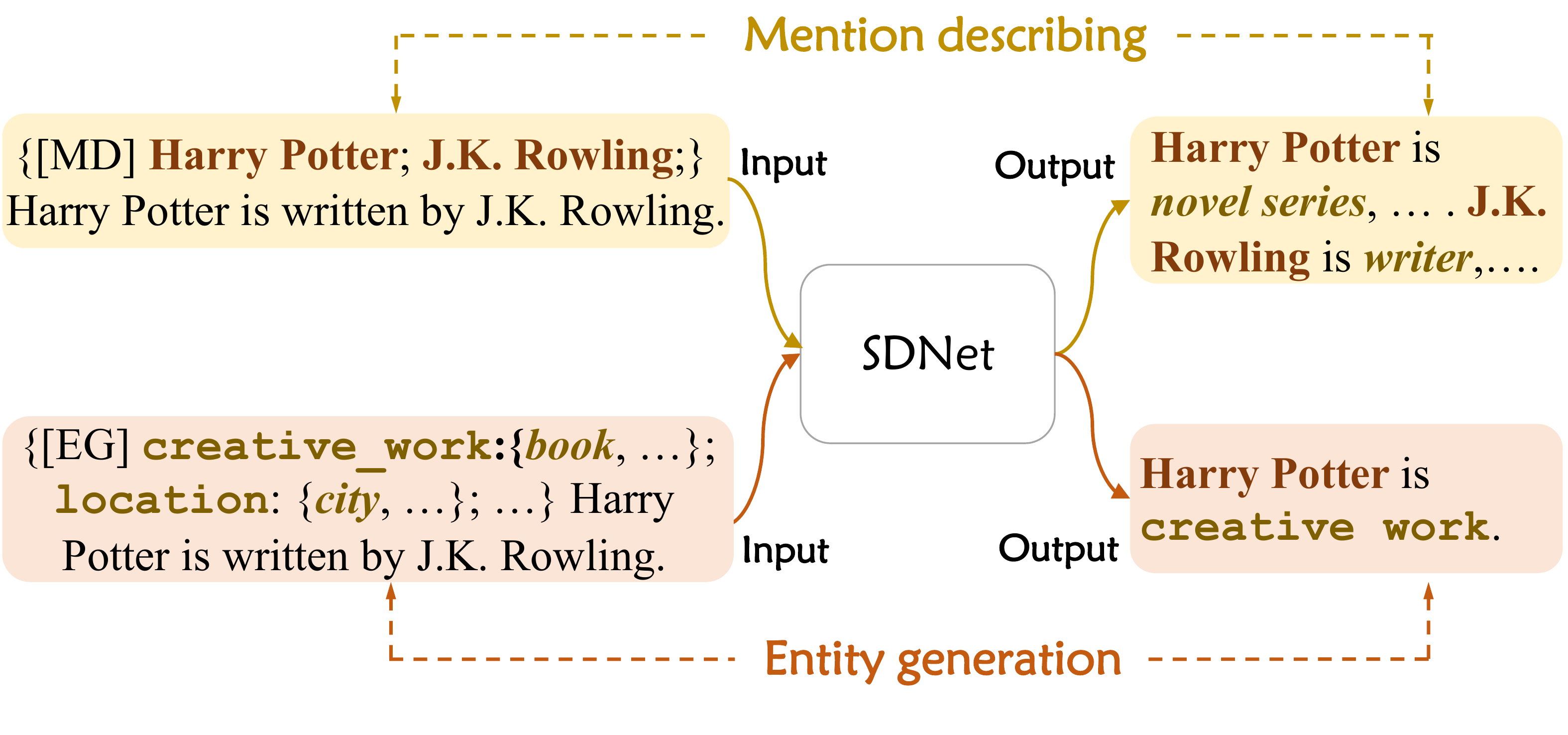}
\caption{Examples of input and output of mention describing and entity generation.}
\label{Fig.ioexample}
\end{figure}

SDNet is a Seq2Seq network that can perform two generation tasks: mention describing and entity generation. Mention describing is to generate the concept descriptions of mentions and entity generation is to adaptively generate entity mentions. To guide the above two processes, SDNet uses different task prompts $\mathbf{P}$ and generates different outputs $\mathcal{Y}$. Figure~\ref{Fig.ioexample} shows their examples. For mention describing, the prompt contains a task descriptor \texttt{[MD]}, and the target entity mentions. For entity recognition, the prompt contains a task descriptor \texttt{[EG]}, and a list of target novel types and their corresponding descriptions. Taking prompt $\mathbf{P}$ and sentence $X$ as input, SDNet will generate a sequence $\mathcal{Y}$ which contains the mention describing or entity generation results. The above two processes can be viewed as \textbf{symmetrical processes}: one is to capture concept semantics of given entities, the other is to identify entities containing the specific concepts.

Specifically, SDNet first concatenates prompt $\mathbf{P}$ and sentence $\mathbf{X}$ into a sequence $\mathcal{I} =\mathbf{P}\oplus \mathbf{X}$ and then fed $\mathcal{I}$ into an encoder to obtain the hidden state representation $\mathbf{\mathcal{H}}$:
$$
\mathbf{\mathcal{H}} = {\rm Encoder}(\mathcal{I}).
$$
Then $\mathbf{\mathcal{H}}$ will be fed into a decoder, and the decoder will sequentially generate a sequence $\mathcal{Y}$. At time step t, the probability $\mathbf{p}_t$ of generating tokens in vocabulary is calculated by:
$$
\mathbf{p}_t= {\rm Decoder}(\mathbf{\mathcal{H}}, \mathcal{Y}_{<t}).
$$

We use the greedy decoding here and therefore the word in the target vocabulary with maximum value in $\mathbf{p}_t$ is generated until $[EOS]$ is generated.

By modeling different tasks in a single model, the generation is controllable, learning is uniform, and the model can be continuously trained.

We can see that, few-shot entity recognition can be effectively performed using the above two generation processes. For entity recognition, we can put the descriptions of target entity types into the prompt, then entities will be adaptively generated through the entity generation process. To construct the entity recognizer of a novel type, we only need its type description, which can be effectively built by summarizing the concept descriptions of their illustrative instances.

\subsection{Entity Recognition via Entity Generation}\label{entity generation}
In SDNet, entity recognition is performed by the entity generation with the given entity generation prompt $\mathbf{P}_\textsc{EG}$ and sentence $X$. Specifically, $\mathbf{P}_\textsc{EG}$ starts with a task descriptor \texttt{[EG]}, and the descriptor is followed by a list of target types and their corresponding descriptions, i.e., $\mathbf{P}_\textsc{EG}$ = $\{\texttt{[EG]}t_1: \{ l^1_1, \ldots, l^{m_1}_1\}; t_2:\{ l^1_2, \ldots, l^{m_2}_2\}; \ldots\}$, where $l^j_i$ is the j-th concept of i-th type $t_i$. Prompt $\mathbf{P}_\textsc{EG}$ and sentence $X$ will be fed to SDNet as Section \ref{overall} described. Then, SDNet will generate text $\mathcal{Y}$ in the format as ``$e_1$ is $t_{y_1}$; \ldots ; $e_n$ is $t_{y_n}.$'', where $t_{y_i}$ is the type of i-th entity $e_i$. Based on the generated text $\mathcal{Y}$, the recognized entities are obtained, i.e, \{<$e_1$, $t_{y_1}$> ... <$e_n$, $t_{y_n}$>\}.

We can see that, in SDNet, the entity generation process can be controlled on-the-fly using different prompts. For example, given a sentence ``Harry Potter is written by J.K. Rowling.'', if we want to identify entity of \texttt{person} type , put \{\texttt{[EG]} \texttt{person}: \{\textit{actor}, \textit{writer}\}\} to $\mathbf{P}_\textsc{EG}$, SDNet will generate ``J.K. Rowling is person'', while if we want to identify entity of \texttt{creative\_work} type, put \{\texttt{[EG]} \texttt{creative\_work}: \{\textit{book}, \textit{music}\}\} to $\mathbf{P}_\textsc{EG}$, SDNet will generate ``Harry Potter is creative\_work''.

\subsection{Type Description Construction via Mention Describing}
SDNet is controlled on-the-fly to generate different types of entities by introducing different corresponding type descriptions to $\mathbf{P}_\textsc{EG}$. For example, the description \{\textit{actor}, \textit{doctor}, ...\} for  and  the description is \{\textit{city}, \textit{state}, ...\} for \texttt{location}.

To build the type description for novel types with several illustrative examples, SDNet first obtains the concept description of each mention in illustrative examples via mention describing. Then the type description of each type is constructed by summarizing all the concept descriptions of its illustrative examples. In the following, we describe them in detail.

\paragraph{Mention Describing.}\label{mention describing}
In SDNet, mention describing is a generation process, whose input is mention describing prompt $\mathbf{P}_\textsc{MD}$ and an illustrative instance $X$. Specifically, given an illustrative example $X$ which contains entity mentions $\{e_1,e_2,\ldots\}$ of novel types, $\mathbf{P}_\textsc{MD}$ starts with a task descriptor \texttt{[MD]}, and the descriptor is followed by target entity mentions. i.e., $\mathbf{P}_\textsc{MD}=\{\texttt{[MD]} e_1; e_2; \ldots\}$. Prompt $\mathbf{P}_\textsc{MD}$ and sentence $X$ will be fed to SDNet as Section \ref{overall} described. And then SDNet will generate the text $\mathcal{Y}$ in the format as ``$e_1$ is $l^1_{1}$, \ldots, $l^{n_1}_1$; $e_2$ is $l^1_{2}$, \ldots, $l^{n_2}_2$; \ldots'', where $l^j_i$ is the j-th concept for the i-th entity mention. The concept set $\{l^1_i, l^2_i, \ldots, l^{n_i}_i\}$ will be considered as the semantic concepts reflected by entity mention $e_i$.

\paragraph{Type Description Construction.} SDNet then summarizes the generated concepts to describe the precise semantics of specific novel types. Specifically, all concept descriptions of mentions with the same type $t$ will be fused to $C$ and regarded as the description of type $t$. And the type descriptions $\mathcal{M} = \{(t,C)\}$ are constructed. Then the constructed type descriptions are incorporated to $\mathbf{P}_\textsc{EG}$ to guide entity generation.

\paragraph{Filtering Strategy.}
Because of the diversified downstream novel types, SDNet may not have sufficient knowledge for describing some of these types, and therefore forcing SDNet to describe them can result in the inaccurate descriptions. To resolve this problem, we introduce a filtering strategy to make SDNet able to reject generating unreliable descriptions. Specifically, SDNet is trained to generate \texttt{other} as the concept description for those uncertain instances. Given a novel type and a few illustrative instances, we will count the frequency of \texttt{other} in the concept descriptions from these instances. If the frequency of generating \texttt{other} on illustrative instances is greater than 0.5, we will remove the type description, and directly use the type name as $\mathbf{P}_\textsc{EG}$. We will describe how SDNet learns the filtering strategy in Section~\ref{sec:pretrain}.

\section{Learning}
In this section, we first describe how to pre-train SDNet using large-scale external data, so that the common NER ability can be captured through the mention describing ability and the entity generation ability. Then we describe how to quickly adapt and transfer NER knowledge via fine-tuning. Figure~\ref{Fig.framework} shows the two processes and we describe them as follows.

\subsection{SDNet Pre-training}\label{sec:pretrain}
In SDNet, the NER ability consists of mention describing ability and entity generation ability, which can be effectively pre-trained by constructing corresponding datasets. This paper constructs datasets and pre-trains SDNet using the easily available and large-scale Wikipedia and Wikidata data.

\paragraph{Entity Mention Collection.}
For SDNet pre-training, we need to collect <$e, T, X$> triples, where $e$ is entity mention, $T$ is entity types and $X$ is sentence, such as <J.K. Rowling; person, writer, ...; J.K. Rowling writes ...>. To this end, we use the 20210401 version of Wikipedia and Wikidata dump and collect triples by aligning facts in Wikidata and documents in Wikipedia and process as follows. 1) Firstly, we construct an entity type dictionary from Wikidata. We regard each item in Wikidata as an entity and use the “instance of”, “subclass of” and “occupation” property values as its corresponding entity types. To learn general NER knowledge, we use all entity types except whose instances are < 5. For the types whose names are longer than 3 tokens, we use their head words as the final type for simplicity, e.g., “state award of the Republic of Moldova” is converted to “state award”. In this way, we obtain a collection $\mathbb{T}$ of 31K types which can serve as a solid foundation for universal NER. 2) Secondly, we collect the mentions of each entity using its anchor texts in Wikipedia and the top 3 frequent noun phrase occurrences of its entry page~\citep{DBLP:conf/acl/LiJW10}. Then for each mention, we identify its entity types by linking it to its Wikidata item's types. If its Wikidata item doesn't have a type, we assign its type as \texttt{other}. For each Wikipedia page, we split the text to sentences\footnote{nltk.tokenize.punkt} and filter out sentences that have no entities. Finally, we construct a training dataset containing 56M instances.

\paragraph{Type Description Building.}
To pre-train SDNet, we need the concept descriptions $\mathcal{M}^P=\{(t_i,C_i)\}$, where $t_i \in \mathbb{T}$, $C_i$ is the related concepts of type $t_i$. This paper uses the collected entity types above as concepts, and builds the type description as follows. Given an entity type, we collect all its co-occurring entity types as its describing concepts. For example, \texttt{Person} can be described as \{\textit{businessman}, \textit{CEO}, \textit{musician}, \textit{musician}...\} by collecting the types of ``Steve Jobs'': \{\texttt{person}, \texttt{businessman}, \texttt{CEO}\} and ``Beethoven'': \{\texttt{person}, \texttt{musician}, \texttt{pianist}\}. In this way, for each entity type we have a describing concept set. Because some entity types have a very large describing concept set, we randomly sample no more than N (10 in this paper) concepts during pre-training for efficiency.
\begin{table*}[th!]
\centering
\setlength{\belowcaptionskip}{-0.2cm}
\resizebox{1.0\textwidth}{!}{
\begin{tabular}{@{}l|lccccccccc@{}}
\toprule
  &
   &
  \textbf{CoNLL} &
  \textbf{WNUT} &
  \textbf{Res} &
  \textbf{Movie1} &
  \textbf{Movie2} &
  \textbf{Re3d} &
  \textbf{I2B2} &
  \textbf{Onto} &
  \textbf{AVE} \\ \midrule
\multirow{6}{*}{\textbf{Baselines}} &
  RoBERTa~\cite{huang} &
  53.5 &
  25.7 &
  48.7 &
  51.3 &
  / &
  / &
  36.0 &
  57.7 &
  / \\
 & RoBERTa-DS~\cite{huang}* & 61.4          & 34.2 & 49.1 & 53.1 & /    & /    & 38.5          & 68.8          & /    \\
 & Proto~\cite{huang}         & 58.4          & 29.5 & 44.1 & 38.0 & /    & /    & 32.0          & 53.3          & /    \\
 & Proto-DS~\cite{huang}*     & 60.9          & 35.9 & 48.4 & 43.8 & /    & /    & 36.6          & 57.0          & /    \\
 & spanNER~\cite{wang}        & 71.1          & 25.8 & 49.1 & /    & 65.4 & /    & /             & 67.3          & /    \\
 & spanNER-DS~\cite{wang}*  & \textbf{75.6} & 38.5 & 51.2 & /    & 67.8 & /    & /             & \textbf{71.6} & /    \\ \midrule
\multirow{4}{*}{\textbf{\begin{tabular}[c]{@{}l@{}}Baselines\\ {[}in-house{]}\end{tabular}}} &
  Bert-base &
  58.6 &
  23.2 &
  47.6 &
  52.4 &
  66.3 &
  57.0 &
  47.6 &
  61.1 &
  51.7 \\
 & T5-base               & 60.0          & 36.6 & 59.4 & 57.9 & 69.9 & 57.1 & 39.9          & 62.0          & 55.3 \\
 & T5-base-prompt      & 55.4          & 34.2 & 58.4 & 58.7 & 67.1 & 60.7 & 61.8 & 59.8          & 57.0 \\
 & T5-base-DS          & 68.2          & 34.9 & 59.7 & 58.4 & 70.8 & 56.0 & 34.1          & 58.8          & 55.1 \\ \midrule
\textbf{Ours} &
  SDNet &
  71.4 &
  \textbf{44.1} &
  \textbf{60.7} &
  \textbf{61.3} &
  \textbf{72.6} &
  \textbf{65.4} &
  \textbf{64.3} &
  71.0 &
  \textbf{63.8} \\ \bottomrule
\end{tabular}
}
\caption{Micro-F1 scores on 8 datasets in 5-shot setting. * means these approaches use external distant supervision datasets to pre-train model different from SDNet. AVE are the average scores of these datasets.}
\label{tab:main results}
\end{table*}

\paragraph{Pre-training via Mention Describing and Entity Generation.}
Given a sentence $X$ with its mention-type tuples $\{(e_i,T_i)|e_i\in E,T_i\subset \mathbb{T}\}$, where $T_i=\{t_i^1, ..., t_i^{n_i}\}$ is the set of types of i-th entity mention $e_i$, $t^j_i$ is the j-th type of the $e_i$, $E=\{e_1,e_2,...\}$ is the set of entity mentions contained in $X$. Then we construct type descriptions, and transform these triples to pre-training instances. Specifically, for mention describing, some target entity mentions $E^{'}$ are sampled from $E$ to put into prompt $\mathbf{P}_\textsc{MD}$. Then SDNet will take $\mathbf{P}_\textsc{MD}$ and $X$ to generate the corresponding types of sampled mentions $E^{'}$ as described in Section \ref{mention describing}. For entity generation, positive type $T_p$ and negative type $T_n$ are sampled to construct the target-sampled type set $T^{'}= T_p \cup T_n$, where $T_p \subset T_1 \cup T_i ... \cup T_k$, $T_n \subset \mathbb{T} \setminus \{T_1 \cup T_i ... \cup T_k\}$. Next, the type set $T^{'}$ and their sampled concept description will be put into prompt $\mathbf{P}_\textsc{EG}$. Then SDNet will take prompt $\mathbf{P}_\textsc{EG}$ and sentence $X$ to generate the sequence as described in Section \ref{entity generation}.

For each instance, SDNet generates two kinds of sequences: $\widetilde{\mathcal{Y}_m^p}$ for mention describing, and $\widetilde{\mathcal{Y}_e^p}$ for entity generation. We use cross-entropy (CE) loss to train SDNet:
\begin{align}
\mathcal{L}_{p} &= \textup{CE}(\widetilde{\mathcal{Y}_m^p}, \mathcal{Y}_m^p) + \textup{CE}(\widetilde{\mathcal{Y}_e^p}, \mathcal{Y}_e^p) \label{pretrain_loss}
\end{align}

Note that when constructing the target generation sequence $\mathcal{Y}_e^p$, the order of mentions depends on the order they appear in the original text.

\subsection{Entity Recognition Fine-tuning}
As described above, SDNet can directly recognize entities using manually designed type descriptions. But SDNet can also automatically build type descriptions using illustrative instances and be further improved by fine-tuning. Specifically, given annotated <$e, T, X$> instances, we first construct the descriptions of different types, next build an entity generation prompt $\mathbf{P}_\textsc{EG}$, then generate sequence $\widetilde{\mathcal{Y}_n^f}$. We fine-tune SDNet by optimizing:
\begin{align}
\mathcal{L}_{f} &= \textup{CE}(\widetilde{\mathcal{Y}_n^f}, \mathcal{Y}_n^f) \label{finetune_loss} 
\end{align}

We can see that, by fine-tuning SDNet, the entity generation process can better capture the associations between mentions and entity types.

\section{Experiments}
\subsection{Settings}
\paragraph{Datasets.}
Following previous studies, we use 8  benchmarks from different domains: 1) CoNLL2003~\citep{conll} ; 2) WNUT17 \citep{wnut}; 3) Re3d~\citep{re3d}; 4) MIT corpus~\citep{mit-res,mit-movie2} includes three datasets: Res, Movie1(trivial10k13 version) and Movie2; 5) I2B2 \citep{i2b2}; 6) OntoNotes5 \citep{ontonotes}. Appendix shows detailed statistics of these datasets.

\paragraph{Evaluation.}
We conduct main experiments on 5-shot setting as previous work~\citep{huang,wang}, and also ranging the shot size from 5 to 100, as well as full shot for further analysis. For k-shot setting, we sample k instances for each entity type from training set as support set to fine-tune models. Specifically, all pre-trained models are trained 300k steps, all datasets are fine-trained 50 epochs and more hyperparameters are shown in Appendix. The performance is evaluated by micro-F1 on test set, and a predicted entity is correct if its entity type and offsets both match the golden entity. To obtain the offset of each mention, we extract entity mentions and their types from the generated sentence, and locate them in the original sentence. And if they are repeated, we match them in order, that is, the i-th same mention in the generated sentence will be matched to the i-th same utterances in the original sentence. We run 10 times for each dataset and report the average F1 score as \citet{huang} and \citet{wang} did.

\begin{table*}[]
\centering
\setlength{\belowcaptionskip}{-0.2cm}
\resizebox{0.9\textwidth}{!}{
\begin{tabular}{@{}l|ccc|ccc|ccc|ccc@{}}
\toprule
 &
  \multicolumn{3}{c|}{\textbf{WNUT}} &
  \multicolumn{3}{c|}{\textbf{Re3d}} &
  \multicolumn{3}{c|}{\textbf{Res}} &
  \multicolumn{3}{c}{\textbf{Movie1}} \\ \midrule
 &
  P &
  R &
  F &
  P &
  R &
  F &
  P &
  R &
  F &
  P &
  R &
  F \\
\textbf{SDNet} &
  \textbf{54.78} &
  37.08 &
  \textbf{44.06} &
  \textbf{63.67} &
  \textbf{67.22} &
  \textbf{65.39} &
  \textbf{63.99} &
  \textbf{57.88} &
  \textbf{60.74} &
  \textbf{63.54} &
  59.30 &
  \textbf{61.33} \\
\multicolumn{1}{r|}{w/o desp} &
  48.78 &
  \textbf{39.51} &
  43.54 &
  62.15 &
  65.87 &
  63.95 &
  62.60 &
  57.44 &
  59.88 &
  62.93 &
  \textbf{59.61} &
  61.22 \\
\multicolumn{1}{r|}{w/o joint} &
  50.68 &
  37.46 &
  42.96 &
  62.99 &
  65.01 &
  63.97 &
  63.15 &
  57.23 &
  60.01 &
  62.71 &
  58.64 &
  60.60 \\ 
\multicolumn{1}{r|}{w/o filter} &
  53.57 &
  35.01 &
  42.23 &
  63.49 &
  66.63 &
  65.00 &
  63.31 &
  57.40 &
  60.17 &
  62.99 &
  59.07 &
  60.96 \\ \bottomrule
\end{tabular}
}
\caption{Ablation experiments. SDNet is the full model, SDNet w/o filter is the same model fine-tuned without filtering strategy, SDNet w/o desp is the same model pre-trained and fine-tuned without description, and SDNet w/o joint is two models trained to perform mention description and entity generation separately.}
\label{tab:ablation}
\end{table*}

\paragraph{Baselines.} We compare with following baselines: 
To evaluate the effect of pre-training for few-shot NER, we compare with baselines without NER-specific pre-training: 1) \textbf{BERT-base}, a traditional sequential BIO-based NER tagger \citep{wang} using pre-trained bert-base-uncased model. 2) \textbf{T5-base}, a generation-based NER baseline which uses the same generation format as SDNet but only using original t5-base model for generation. 3) \textbf{T5-base-prompt}, the prompt-extended version of T5-base which use entity types as prompt.
To compare the effect of different knowledge transfer ways, we construct a distant supervision based baseline: 4) \textbf{T5-base-DS}, we further pre-train T5-base using the dataset collected in Section~\ref{sec:pretrain} as distantly supervised dataset. 
We also compare with several recent few-shot NER methods: 5) RoBERTa-based few-shot classifier \textbf{RoBERTa} and its distantly-supervised pre-trained version \textbf{RoBERTa-DS} \citep{huang}. 6) Prototypical network based RoBERTa model \textbf{Proto} and its distantly supervised pre-training version \textbf{Proto-DS} \citep{huang}. 7) MRC model \textbf{SpanNER} which needs to design the description for each label and its distantly supervised pre-training version \textbf{SpanNER-DS} \citep{wang}. Notice that these methods mostly only focus on task-specific entity types, by contrast, this paper focuses on building a general few-shot NER model which can recognize entities universally.

\subsection{Main Results}

Table~\ref{tab:main results} shows the performances of SDNet and all baselines on 8 datasets. We can see that:

1) \textbf{By universally modeling and pre-training NER knowledge in a generation architecture, the self-describing network can effectively handle few-shot NER.} Compared with previous baselines, SDNet achieves competitive performance on all 8 datasets (new state-of-the-art on 6 datasets), and its performance is robust on different datasets.

2) \textbf{Due to the limited information problem, transferring external knowledge to few-shot NER models are critical.} Compared with BERT-base, T5-base, and T5-base-prompt, SDNet achieves 24\%/16\%/11\% F1 improvements, which verified that SDNet provides a universal knowledge-enhanced foundation for NER and can adaptively transfer universal knowledge to enhance novel type recognition.

3) \textbf{Due to the knowledge mismatch, it is challenging to transfer external knowledge effectively to novel downstream types.} Using the same external knowledge sources, SDNet can achieve a 16\% F1 improvement than T5-base-DS. We believe it is due to the noise, partially-labeled and heterogeneous problems in the external knowledge sources, and SDNet can effectively address these issues.

\subsection{Effects of Shot Size}

\begin{figure}[tbh!] 
\setlength{\belowcaptionskip}{-0.2cm}
\centering 
\includegraphics[width=0.48\textwidth]{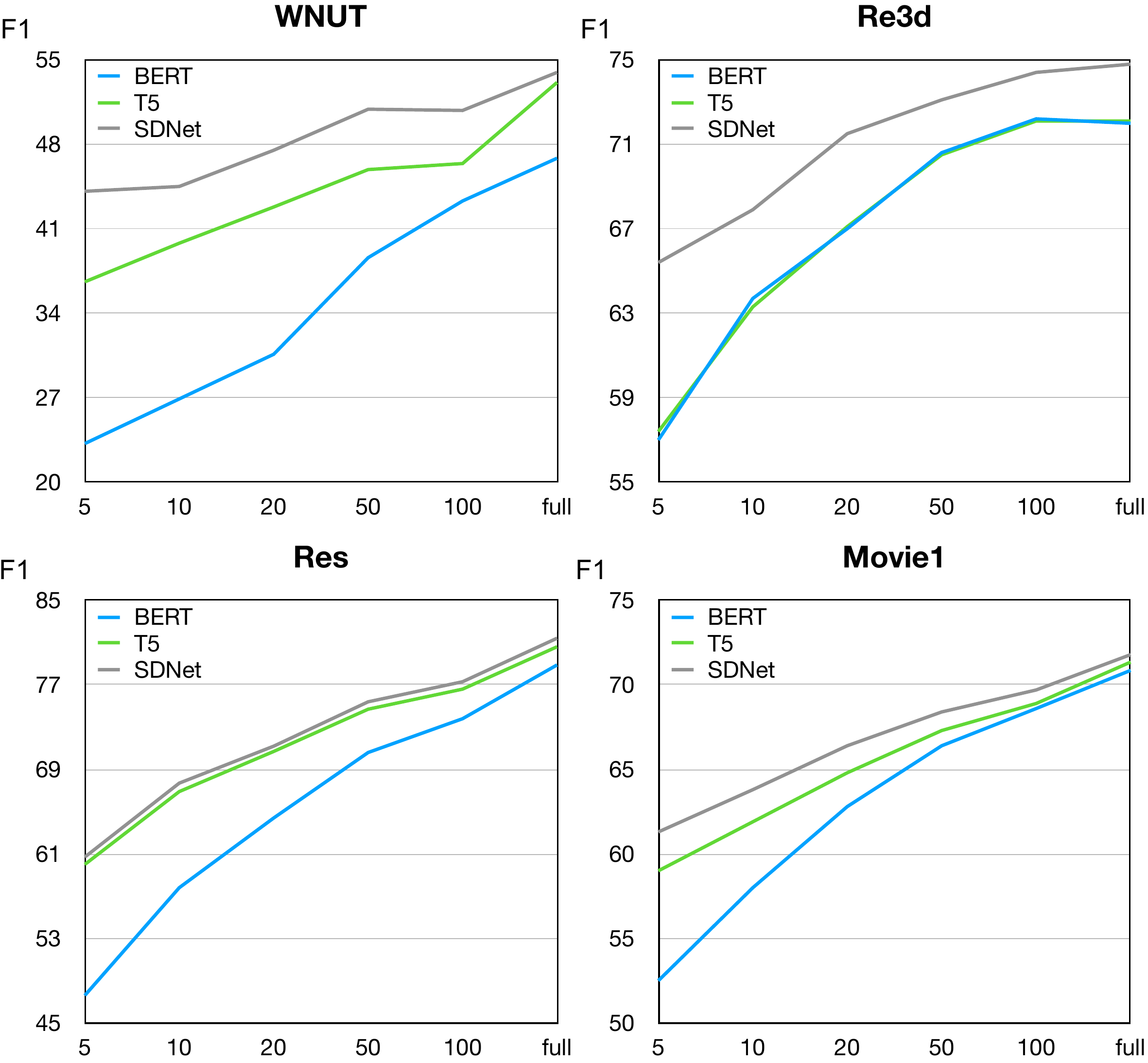}   
\caption{Performances of BERT, T5, SDNet with k-shot samples on WNUT, Re3d, Res, Movie1 dataset.}
\label{Fig.k-shot}
\end{figure}

To verify the performance of SDNet under different shot settings, we compare the performance of BERT, T5, and SDNet with k-shot samples where k ranges from 5 to 100. 
From Figure~\ref{Fig.k-shot} we can see that 1) SDNet can achieve better performance under all different shot settings. Furthermore, the improvements are more significant on low shot settings, which verified the intuitions behind SDNet; 
2) Generation-based models usually achieve better performance than classifier-based BERT model. We believe this is because generation-based model can more efficiently capture the semantics of types by leveraging the label utterances, and therefore can achieve much better performance, especially in low-shot settings.
3) SDNet significantly outperforms T5 on almost all datasets except \textit{Res}. This shows the effectiveness of the proposed self-describing mechanism. For \textit{Res}, we find that the main reason why T5 can achieve close performance to SDNet is the huge domain shifting between \textit{Res} and Wikipedia. Such domain shifting makes SDNet frequently generate \textit{other} for type descriptions, and therefore SDNet degrades to T5 in many cases. However, SDNet can still perform better than T5 on \textit{Res}, which verifies the robustness of the proposed type description and the filtering strategy.

\subsection{Ablation Study}

To analyze the effectiveness of type description, multi-task modeling and type description filtering, we conduct following ablation experiments: 
1) SDNet w/o desp: we directly use entity type as prompt, without the universal concept description, e.g. \{\texttt{[EG]} \texttt{person}; \texttt{location}; \dots\} ; 2) SDNet w/o joint: we split SDNet into two individual generation network, one for mention description, the other for entity generation, and trained them using same resources as SDNet; 3) SDNet w/o filter: we use all the generated concept descriptions with no filtering strategy. From Table~\ref{tab:ablation} we can see that:

1) \textbf{Type description is critical for SDNet to transfer knowledge and capture type semantics.} By removing type description, the F1 of all datasets will decrease. We believe this is because 1) type description provides a common base for knowledge transferring, where all entity types are described using the same set of concepts; 2) the concept descriptions capture the semantics of entity types more accurately and precisely, which can better guide the NER process.

2) \textbf{Joint learning mention describing and entity generation processes in a unified generation network is effective to capture type semantics.} Compared with modeling two tasks separately, SDNet can achieve better performance. We believe this is because the two processes are symmetrical, and they can complement and promote each other.

3) \textbf{Filtering strategy can effectively alleviate the transferring of mismatched knowledge.} Removing the filtering strategy will undermine the performance on all 4 datasets. We believe this is because there exist some instances that can not be described based on the pre-trained SDNet knowledge. As a result, introducing filtering strategy can effectively prevent the mistaken knowledge transferring to these instances.

\subsection{Zero-shot NER with Manual Description}

In this section, we adapt SDNet to zero-shot setting, to investigate whether SDNet can achieve promising zero-shot performance without any illustrative instances. To this end, we conduct an experiment on WNUT by introducing manually created concepts as type descriptions based on annotation guideline, and the designed descriptions are shown in Appendix. Then we compare with the baseline without using type description, to see the effectiveness of the descriptions and whether SDNet can well-adapted to manually created descriptions.

From Table~\ref{tab:label analysis}, we can see that SDNet can benefit from manual description significantly. Compared with SDNet without description, incorporating manual description can improve zero-shot performance on the majority of types. Furthermore, SDNet with manual description on zero-shot setting can achieve comparable performance with few-shot settings in many entity types. This demonstrates that type description is an effective way for model to capture the semantic of novel types, which verifies the intuition of SDNet.

\begin{table}[]
\centering
\setlength{\belowcaptionskip}{-0cm}
\resizebox{0.4\textwidth}{!}{\begin{tabular}{@{}l|cc|c@{}}
\toprule
\textbf{Types} & \multicolumn{1}{l}{\textbf{w/o desp}} & \textbf{Artifact} & \textbf{Few-shot} \\ \midrule
person         & 51.29                                 & 46.27           & 59.80             \\
corporation    & 31.30                                  & 34.43           & 33.21             \\
location       & 46.15                                 & 50.36           & 52.64             \\
creative-work  & 12.29                                 & 14.38           & 27.69             \\
group          & 19.93                                 & 24.45           & 24.10             \\
product        & 19.10                                  & 23.08           & 23.63             \\ \bottomrule
\end{tabular}}
\caption{F1 score of each label in WNUT under zero-shot setting. \textbf{w/o desp} is the model pre-trained without description. \textbf{Artifact} is SDNet with manually designed concept description based on guideline.}
\label{tab:label analysis}
\end{table}

\subsection{Effect of Entity Generation Prompt}
\begin{table}[]
\setlength{\belowcaptionskip}{-0.4cm}
\centering
\resizebox{0.49\textwidth}{!}{
\begin{tabular}{@{}l|l@{}}
\toprule
\textbf{Text}    & [Chris Hill]$_{\text{person}}$ was in [China]$_{\text{GPE}}$ [a few days ago]$_{\text{date}}$. \\ \toprule
\textbf{Input1} & \begin{tabular}[c]{@{}l@{}}\{ {[}EG{]} \ GPE: \{state, country, city, democracy, republic,  \\
community\}; date: \{\}; \} Chris Hill was in China ...\end{tabular} \\\midrule
\textbf{Output1} & China is GPE. a few days ago is date.   \\ \toprule
\textbf{Input2} & \begin{tabular}[c]{@{}l@{}}\{ {[}EG{]} person: \{politician, actor, lawyer\};  organization:  \\ \{business, company\}; \} Chris Hill was in China ...\end{tabular}                      \\\midrule
\textbf{Output2} & Chris Hill is person.                   \\ \bottomrule
\end{tabular}
}
\caption{Examples of the outputs of entity generation process. SDNet is fine-tuned on OntoNotes and the type description is automatically generated by SDNet via mention describing.}
\label{tab:case}
\end{table}
Table~\ref{tab:case} shows that by putting different types and its corresponding type descriptions to prompt, SDNet can generate different outputs according to the prompt. This verifies that SDNet can be controlled on-the-fly to generate different types of entities.

\section{Conclusions}
In this paper, we propose Self-describing Networks, a Seq2Seq generation model which can universally describe mentions using concepts, automatically map novel entity types to concepts, and adaptively recognize entities on-demand. A large-scale SDNet model is pre-trained to provide universal knowledge for downstream NER tasks. Experiments on 8 datasets show that SDNet is effective and robust. For future work, we will extend self-describing mechanism to other NLP tasks like event extraction~\cite{DBLP:conf/iclr/PaoliniAKMAASXS21, DBLP:conf/acl/0001LXHTL0LC20} and complex NER tasks like nested~\cite{DBLP:conf/acl/LinLHS19} or discontinuous NER~\cite{DBLP:conf/acl/YanGDGZQ20}.

\section{Acknowledgments}
We thank all reviewers for their valuable comments. This work is supported by the National Key Research and Development Program of China (No. 2020AAA0106400), the National Natural Science Foundation of China under Grants no. U1936207, 62122077 and 62106251, and the Project of the Chinese Language Committee under Grant no. YB2003C002.

\section{Ethics Consideration}
This paper has no particular ethic consideration.

\bibliographystyle{acl_natbib}
\bibliography{few-shotNER.bib, dataset.bib, anthology.bib, custom.bib}

\appendix

\newpage

\section{Experimental Details}
\paragraph{Detailed Hyperparameters}\label{sec:hyperparameters}
\begin{table}[ht!]
\setlength{\belowcaptionskip}{-0.4cm}
\centering
\resizebox{0.35\textwidth}{!}{
\begin{tabular}{c|c|c}
\toprule
&Pre-train&Fine-tune\\\midrule
batch size&16&4\\
Learning rate&5e-5&1e-4\\
Optimizer&AdamW&AdamW\\
schedule&-&linear schedule\\
warmup rate&-&6\%\\\bottomrule
\end{tabular}}
\caption{Hyperparameter settings.}
\label{tab:hyper}
\end{table}
SDNet is initialized with T5-base. Table~\ref{tab:hyper} shows the hyperparameters of SDNet. When fine-tuning, we set the same hyperparameters for T5-base, T5-base-prompt and T5-base-DS as for SDNet. For Bert-base, the learning rate is 2e-5 and batch size is 8.
\paragraph{Dataset Analysis}\label{sec:experiment_data}
Table~\ref{tab:dataset} shows the statistics of the dataset we use.
\paragraph{Manually Designed Type Descriptions} Table~\ref{tab:label description case} shows the manually designed type descriptions for WNUT.
\begin{table}[ht!]
\centering
\resizebox{0.35\textwidth}{!}{
\begin{tabular}{c|c|c|c}
\toprule
Dataset & Domain &\#Types & \#Test\\\midrule
WNUT &Social Media &6 &1287\\
CoNLL &News &4 &3453\\
re3d &Defense &10 &200\\
Res &Review &8 &1521\\
Moive1 &Review &12 &1953\\
Movie2 &Review &12 &2443\\
I2B2 &Medical &23 &43697\\
Onto &General &18 &8262
\\\bottomrule
\end{tabular}}
\caption{Statistics of 8 public datasets.}
\label{tab:dataset}
\end{table}

\begin{table}[ht!]
\centering
\setlength{\belowcaptionskip}{-0.4cm}
\resizebox{0.49\textwidth}{!}{
\begin{tabular}{|l|l|}
\hline
\textbf{person}        & \begin{tabular}[c]{@{}l@{}}writer, entrepreneur, association football player, actor, \\  businessperson, baseball player, politician\end{tabular} \\ \hline
\textbf{corporation}   & \begin{tabular}[c]{@{}l@{}}digital media, website, organization, trademark, \\ entrepreneur, airline, social media\end{tabular}                  \\ \hline
\textbf{location}      & port, park, city, country, road, province, state, mountain                                                                                       \\ \hline
\textbf{creative-work} & television program, audiovisual work, album, release, film                                                                                       \\ \hline
\textbf{group}         & group, band, basketball team, football club, sports team                                                                                         \\ \hline
\textbf{product}       & \begin{tabular}[c]{@{}l@{}}medication, chemical compound, electronic game, \\ video game, smartphone model, chemical substance\end{tabular}      \\ \hline
\end{tabular}
}
\caption{Manually designed concept descriptions in WNUT.}
\label{tab:label description case}
\end{table}

\end{document}